\documentclass[10pt,twocolumn,letterpaper]{article}

\usepackage{iccv}
\usepackage{times}
\usepackage{epsfig}
\usepackage{graphicx}
\usepackage{amsmath}
\usepackage{amssymb}
\usepackage{eqnarray}
\usepackage{verbatim}
\usepackage{subfig}
\usepackage{multirow}

\newtheorem{theorem}{Hypothesis}

\usepackage[pagebackref=true,breaklinks=true,letterpaper=true,colorlinks,bookmarks=false]{hyperref}

\newcommand{\lonenorm}[1]{{\mid\!\mid\!{#1}\!\mid\!\mid_1}}
\newcommand{\ltwonorm}[1]{{\mid\!\mid\!{#1}\!\mid\!\mid_2}}

\newcommand{\vect}[1]{{\bf {#1}}}

\newcommand{\relu}[1]{{{\mathsf R}}{({{#1}})}}
\newcommand{\dafabs}[1]{{\mid\! {#1} \! \mid}}

\iccvfinalcopy 

\def\iccvPaperID{166} 

\ificcvfinal\fi
\begin{document}

\title{SafetyNet: Detecting and Rejecting Adversarial Examples Robustly}

\author{Jiajun Lu, Theerasit Issaranon, David Forsyth\\
University of Illinois at Urbana Champaign\\
\{jlu23, issaran1, daf\}@illinois.edu
}

\maketitle

\begin{abstract}
We describe a method to produce a network where current methods such as DeepFool have great difficulty producing adversarial samples.
Our construction suggests some insights into how deep networks work.
We provide a reasonable analyses that our construction is difficult to defeat, and 
show experimentally that our method is hard to defeat with both Type I and Type II attacks using several standard networks and datasets.  
This SafetyNet architecture is used to an important and novel application SceneProof, which can reliably detect whether an image is a picture of a real scene or not.  
SceneProof applies to images captured with depth maps (RGBD images) and checks if a pair of image and depth map
is consistent.  It relies on the relative difficulty of producing naturalistic depth maps for images in post processing.
We demonstrate that our SafetyNet is robust to adversarial examples built from currently known attacking approaches.
\end{abstract}

\section{Introduction}

Adversarial examples are images with tiny, imperceptible perturbations that fool a classifier into predicting the wrong labels with high confidence. 
$\vect{x}$ denotes the input to some classifier, which is a {\em natural} example and has label $l$.
A variety of constructions~\cite{goodfellow2014explaining,kurakin2016adversarial,moosavi2016deepfool,papernot2016practical} can generate 
an {\em adversarial} example $\vect{a}(\vect{x})$ to make the classifier label it 
$m \neq l$.  This is interesting, because $\ltwonorm{\vect{a}(\vect{x})-\vect{x}}$ is so small that we 
would expect $\vect{a}(\vect{x})$ to be labelled $l$. 

\begin{figure}[ht]
\centerline{\includegraphics[width=0.5 \textwidth]{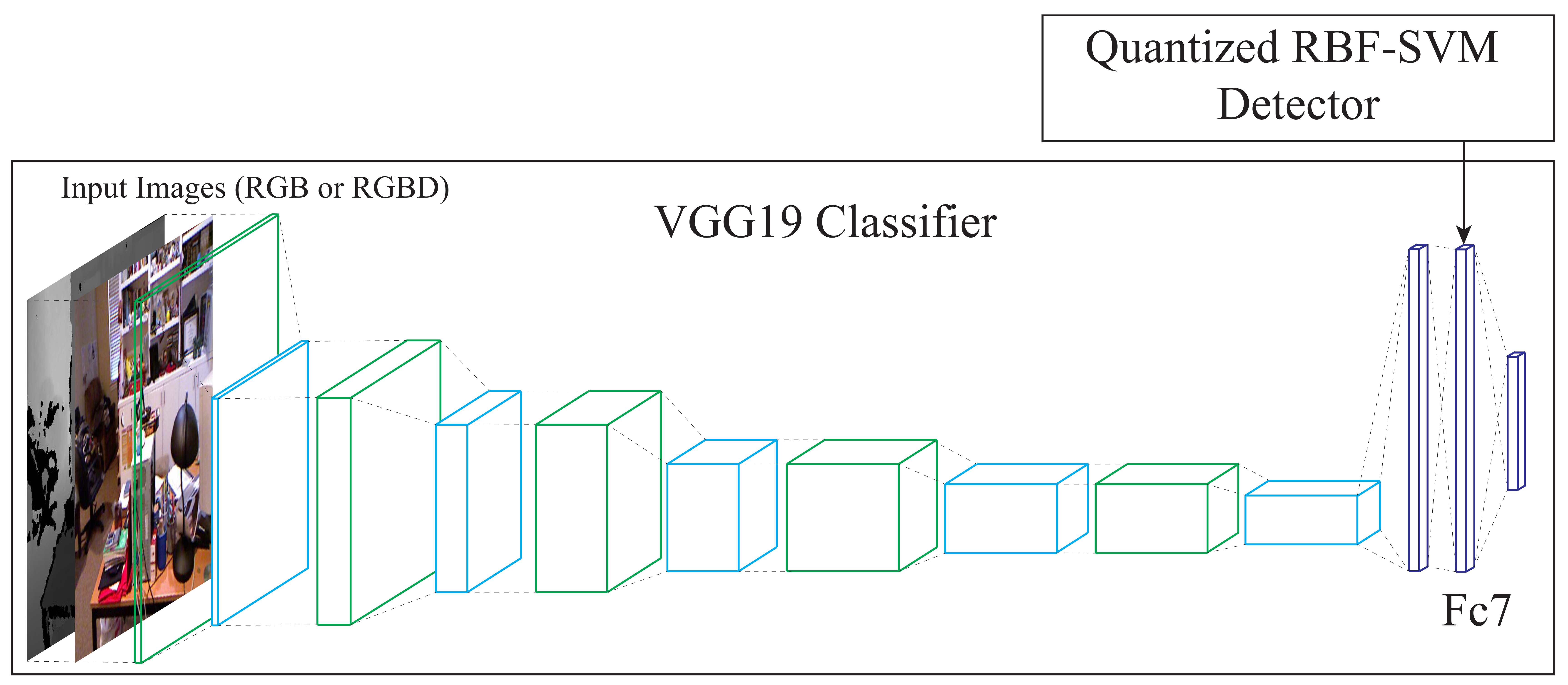}}
  \caption{SafetyNet consists of a conventional classifier (in our
    experiments, either VGG19 or ResNet) with an RBF-SVM that uses discrete codes
computed from late stage ReLUs to detect adversarial examples. We show that (a) SafetyNet detects adversarial examples
reliably, even if they are produced by methods not represented in the detectors' training set and (b) it is very
difficult to produce examples that are both misclassified and slip past SafetyNet's detector.
  \label{fig:detector}}
\end{figure}

Adversarial examples are a persistent problem of classification neural networks, and of many other classification schemes. 
Adversarial examples are easy to construct~\cite{szegedy2013intriguing,nguyen2015deep,carlini2016towards}, and there are even universal adversarial 
perturbations~\cite{moosavi2016universal}.  Adversarial examples are important for practical reasons, because one can
construct physical adversarial examples, suggesting that neural networks in current status are unusable in
some image classification applications (\eg imagine a small physical modification that could reliably get a stop sign classified
as a go faster sign~\cite{papernot2016practical,lu2017no}).   Adversarial examples are important for conceptual reasons too,
because an explanation of why adversarial examples are easy to construct could cast some light on the inner life of
neural networks.  The absence of theory means it is hard to defend against adversarial examples (for example, distillation was proposed as a
defense~\cite{papernot2016distillation}, but was later shown to not work~\cite{carlini2016defensive}).  

Adversarial example constructions (\eg, line search along the gradient~\cite{goodfellow2014explaining}; LBFGS on an
appropriate cost~\cite{szegedy2013intriguing}; DeepFool~\cite{moosavi2016deepfool}) all  rely on the 
gradient of the network, but it is known that using the gradient of another similar network is
sufficient~\cite{papernot2016practical}, so concealing the gradient does not work as a defense for current networks.    
An important puzzle is that networks that generalize very well remain susceptible to adversarial
examples~\cite{szegedy2013intriguing}.  Another important puzzle is that examples  
that are adversarial for one network tend to be adversarial for another as well
~\cite{szegedy2013intriguing,Song2016,papernottransfer}. Some network architectures 
appear to be robust to adversarial examples~\cite{krotov2017}, which still need more empirical verification.
At least some adversarial attacks appear to apply to many distinct networks~\cite{moosavi2016universal}.

We denote the probability distribution of examples by $P(X)$.  At least in the case of vision, $P(X)$ has support on
some complicated subset of the input space, which is known as 
the ``manifold'' of ``real images''.  Nguyen {\em et al.} show how to construct examples that appear to be noise, but
are confidently classified as objects~\cite{nguyenfooled}.  This construction yields $\vect{a}(\vect{x})$ lies outside
the support of $P(X)$, so the classifier's labeling is unreliable because it  has not  seen such examples.   However,
most adversarial examples ``look like'' images to humans, such as figure 5 in~\cite{szegedy2013intriguing}, so they are likely to
lie within the support of $P(X)$.  

One way to build a network that is robust to adversarial examples is to train networks with enhanced training data
(adding adversarial samples~\cite{miyato2015distributional}); this approach faces difficulties, because the dimension of
the images and features in networks means an unreasonable quantity of training data is required. 
Alternatively, we can build a network that detects and rejects an adversarial sample. 
 Metzen {\em et al.} show that, by attaching a detection subnetwork that observes the state of the original
 classification network, one can tell whether it has been presented with an 
adversarial example or not~\cite{metzen2017detecting}. However, because the gradients of their detection subnetwork
are quite well behaved, the joint system can be attacked (Type II attack) easily in both their and our experiments.  
Both their and our experiments also show that their detection subnetwork is easily fooled by adversarial samples produced 
by attacking methods which are not used in detector training process. 

Our method focuses on codes produced by quantizing individual
ReLUs in particular layers of the classification network (``patterns of activation''), and proceed from the hypothesis:
\begin{theorem}
Adversarial attacks work by producing different patterns of activation in late
stage ReLUs to those produced by natural examples.
\label{hyp}
\end{theorem}
These patterns lie outside the family for which the softmax layer would be reliable.
This hypothesis suggests that:  (a) the presence of an adversarial example can be
detected (as in Metzen {\em et al.}~\cite{metzen2017detecting}); (b) such detectors can be made
very difficult to defeat (unlike Metzen {\em et al.}~\cite{metzen2017detecting}; section~\ref{sec:sceneproof}); 
(c). such detectors should be good at generalization for different adversarial attacks 
(unlike Metzen {\em et al.}~\cite{metzen2017detecting}); 
(d) transfer attacks work because an example that generates unfamiliar patterns in one
network tends to generate unfamiliar patterns in other networks too; (e) transfer attacks could be defended as well (section~\ref{sec:sceneproof}).


{\bf Contributions:} 
Section~\ref{sec:safetynet} describes our SafetyNet architecture, which consists of  the original classifier network and
 a detector that rejects adversarial examples.
A \textbf{type I attack} on SafetyNet consists of a standard adversarial example crafted to be (a)
similar to a natural image; (b) misclassified by the original network.  A \textbf{type II attack} consists of an example
that is crafted to be (a) similar to a natural image; (b) misclassified; {\em and} (c) not rejected by SafetyNet.  We
show that SafetyNet is robust to both types of attacks and generalize well. Concealing the gradients is highly effective 
for SafetyNet, and it produces a black box that is strongly resistant to the best attacks we have been able to construct. 
This is in sharp contrast to all other known methods~\cite{papernot2016practical, metzen2017detecting}.

In section~\ref{sec:sceneproof}, we demonstrate SceneProof, a robust and reasonably
effective proof that an image is an image of a real scene (a ``real'' image; contrast a ``fake'' image, which is not an
image of a real scene).   We identify images of real scenes by checking a match between the image and a depth map, which is hard to manipulate. 
We show that SceneProof is (a) accurate and (b) strongly resistant to attacks that try to get manipulated
scenes identified as authentic scenes.

In section~\ref{sec:label}, we propose a model that explains why our approach works, and it also demonstrates that
SafetyNet is difficult to attack in principle.

\section{SafetyNet: Spotting Adversarial Examples}
\label{sec:safetynet}
SafetyNet consists of the original classifier, and an adversary detector which looks at the internal state of the later layers in
the original classifier, as in Figure~\ref{fig:detector}. If the adversary detector declares that an example is adversarial, then
the sample is rejected.

\subsection{Detecting Adversarial Examples}

The adversary detector needs to be hard to attack.  We force an attacker to solve a hard discrete optimization
problem.  For a layer of ReLUs at a high level in the classification network, we quantize each ReLU at some
set of thresholds to generate a  discrete code (binarized code in the case of one threshold).  
Our hypothesis~\ref{hyp} suggests that different code patterns appear for natural examples and
adversarial examples.  We use an adversary detector that compares a code produced at test time with a collection of 
examples, meaning that an attacker must make the network produce a code that is acceptable to the detector (which is
hard; section~\ref{sec:sceneproof}).  The adversary detector in  SafetyNet uses an RBF-SVM on binary or quaternary codes (activation patterns) to find
adversarial examples.  

We denote a code by $\vect{c}$. The RBF-SVM classifies by
\begin{equation}
f(\vect{c}) = \sum_i^N \alpha_i y_i \exp(-||\vect{c} - \vect{c}_i||^2 / 2\sigma^2) + b
\end{equation}
In this objective function, when $\sigma$ is small, the detector produces essentially no gradient unless the attacking code $\vect{c}$ 
is very close to a positive example $\vect{c}_i$.  Our quantization process makes the detector more robust and the gradients 
even harder to get. Experiments show that this form of gradient obfuscation is quite robust, 
and that confusing the detector is very difficult without access to the RBF-SVM, and still difficult even when access is possible.  
Experiments in section~\ref{sec:sceneproof} and theory in section~\ref{sec:label} confirm that the optimization problem is hard.

\subsection{Attacking Methods}

We use the following standard and strong attacks~\cite{carlini2016defensive}, with various choice of
hyper-parameters, to test the robustness of the systems. Each attack searches for a nearby $\vect{a}(\vect{x})$ which 
changes the class of the example and does not create visual artifacts. We use
these methods to produce both type I attack (fool the classifier) and type II attack (fool the classifier {\em and} sneak past the
detector).

\textbf{Fast Sign method: } Goodfellow et al~\cite{goodfellow2014explaining} described this simple method. The applied perturbation is the direction in image space which yields the
highest increase of the linearized cost under $l_{\infty}$ norm. It uses a hyper-parameter $\epsilon$ to govern the 
distance between adversarial and original image. 

\textbf{Iterative methods: } Kurakin et al.~\cite{kurakin2016adversarial} introduced an iteration version of the fast
sign method, by applying it several times with a smaller step size $\alpha$ and clipping all pixels after each iteration
to ensure that results stay in the $\epsilon$ neighborhood of the original image.  We apply two versions of this method, one
where the neighborhood is in $L_\infty$ norm and another where it is in $L_2$ norm.

\textbf{DeepFool method: } Moosavi-Dezfooli et al.~\cite{moosavi2016deepfool} introduced the DeepFool adversary, which
is able to choose which class an example is switched to.  DeepFool iteratively perturbs an image $x_{0}^{adv}$, 
linearizes the classifier around $x_n^{adv}$ and finds the closest class boundary. The minimal step according to the
$l_p$ distance from $x_n^{adv}$ to traverse this class boundary is determined and the resulting point is used as
$x_{n+1}^{adv}$. The algorithm stops once $x_{n+1}^{adv}$ changes the class of the actual classifier.   We use a powerful $L_2$
version of DeepFool.

\textbf{Transfer method: } Papernot et al.~\cite{papernot2016practical} described a way to attack a black-box
network. They generated adversarial samples using another accessible network, which performs the same task, and used
these adversarial samples to attack the black-box network. This strategy has been notably reliable.  

\subsection{Type I Attacks Are Detected}

\begin{figure*}[ht]
\centerline{\includegraphics[width=0.90 \textwidth]{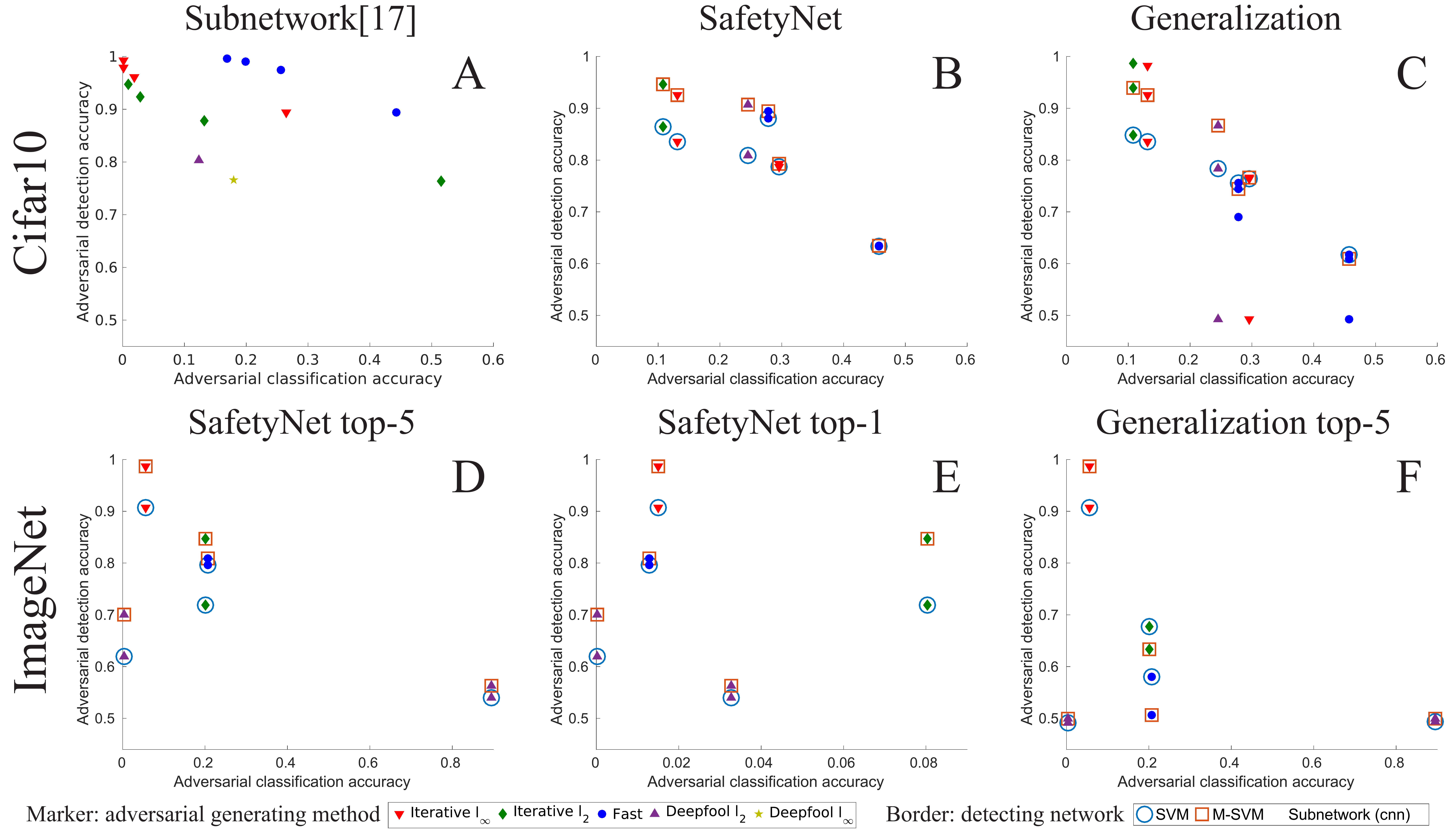}}
\caption{SafetyNet accurately detects adversarial attacks.
To facilitate comparison, we follow the conventions of \protect \cite{metzen2017detecting}, plotting the success of the
adversary (i.e. its ability to fool the classifier; leftward is better) on the horizontal axis and the accuracy of the
detector on the vertical axis (higher is better).  We show results for binary (SVM) and quaternary (M-SVM) codes, and for  a variety of
attacks. {\bf A:} Results for the detection subnetwork on CIFAR-10 from \protect \cite{metzen2017detecting}.  
{\bf B:} Results for SafetyNet on CIFAR-10, where the detector was trained and tested on adversarial samples generated by the 
same attacking method (same setting as {\bf A}).
{\bf C:} Results for SafetyNet {\em and} the detection subnetwork  (cnn) of \protect \cite{metzen2017detecting} on CIFAR-10,
where the detector was trained on $L_\infty$ attack and tested on other attacking methods; SafetyNet
generalizes better than detection subnetwork to different adversarial attacking methods. 
{\bf D:} Results for SafetyNet on ImageNet-1000, where the detector was trained and tested on the same adversarial method.
The classifier is evaluated with top-5 accuracy ({\bf E} is evaluated with top-1 accuracy, note difference in x axis); 
using top-5 accuracy significantly advantages the adversary detector, because forcing an adversarial example to move out of top-5 requires larger changes.
{\bf F:} Results for SafetyNet on ImageNet-1000 (top-5), where the detector was trained on $L_\infty$ attack and tested on other attacking methods; 
SafetyNet has relatively small loss of detection accuracy (compared to {\bf E}).  We cannot compare to 
the detection subnetwork of \protect \cite{metzen2017detecting}, because they do not provide results for ImageNet-1000.
}
  \label{fig:cmp1}
\end{figure*}

\begin{figure*}[h]
\begin{center}
\resizebox{0.90\textwidth}{!}
{
\includegraphics{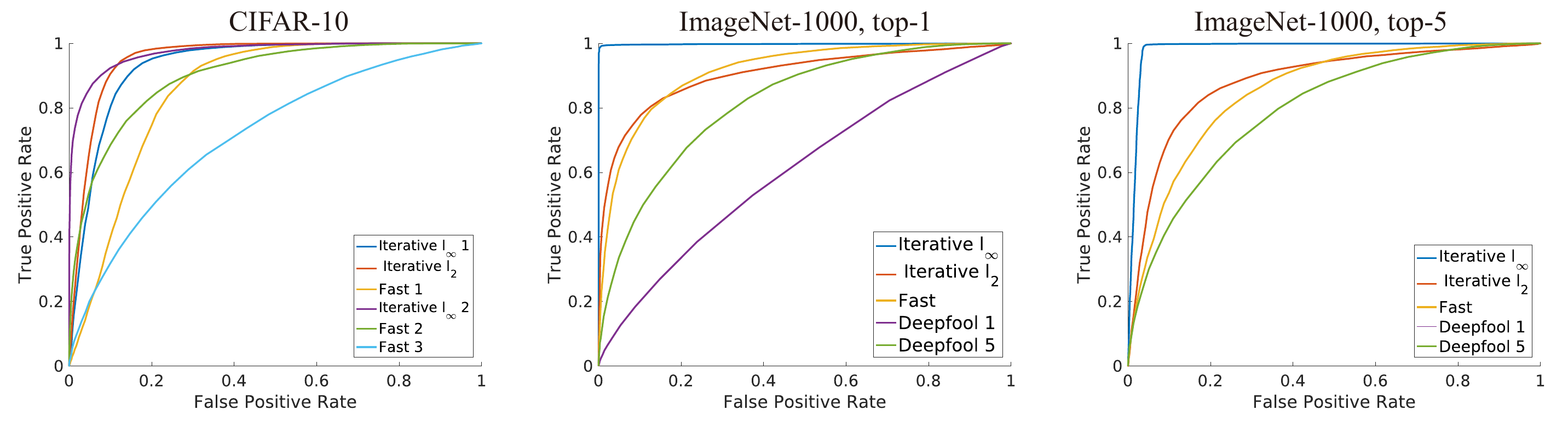}
}
\caption{ROC curve for our adversary detector on various adversaries.  {\bf Left:} CIFAR-10; {\bf center:} ImageNet-1000,
  top-1;  {\bf right:} ImageNet-1000, top-5.  Deepfool-5 is a variant of deepfool that is required to force the
  adversarial example out of the original example's top 5 classes. Deepfool is a strong adversarial attack, and seems to benefit from
  being able to choose the target class from multiple classes.  }
\label{fig:roc}
\vspace{-2ex}
\end{center}
\end{figure*}

{\bf Accuracy:} Our SafetyNet can detect adversarial samples with high accuracy on CIFAR-10~\cite{krizhevsky2009learning} and ImageNet-1000~\cite{deng2009imagenet}. 
For classification networks, we used a 32-layer ResNet~\cite{he2016deep} for CIFAR-10 and a VGG19 network~\cite{Simonyan14c} for ImageNet-1000.
Figures~\ref{fig:cmp1} shows the detection accuracy of our Binarized RBF-SVM detector on 
the x5 layer of ResNet for Cifar10 and on the fc7 layer of VGG19 trained for ImageNet-1000.
Adversarial samples are generated by \textbf{Iterative-L2}, \textbf{Iterative-Linf},
\textbf{DeepFool-L2} and \textbf{FastSign} methods.  Figure~\ref{fig:cmp1} compares 
our RBF-SVM detection results with the detector subnetwork results of~\cite{metzen2017detecting}. 
The RoC for our detector for Cifar-10 and ImageNet-1000 appears in Figure~\ref{fig:roc}. 

Our results show:  When our detector is tested on the same adversary as it is trained on, its performance is similar to
the detector subnetwork~\cite{metzen2017detecting}, even though our detector  works on quantized activation patterns while the detector subnetwork
works on original continuous activation patterns. DeepFool is a strong attack.  Increasing the number of categories in
the problem makes it easier for DeepFool to produce an undetected adversarial 
example, likely because it becomes easier to exploit local classification errors without producing strange ReLU
activations.  If DeepFool is required to produce a label outside the top-5 for the original example, the attack is much
weaker.

{\bf Generalization across attacks:} 
Generally, a detector cannot know at training time what attacks will occur at test time.  We test generalization
across attacks by training a detector on one class of attack, then testing with other classes of attack. 
Figure~\ref{fig:cmp1} shows that our RBF-SVM generalizes across attacks more reliably than a detector subnetwork.
We believe this is because the representation presented
to the RBF-SVM has been aggressively summarized (by quantization), so that the classifier is not distracted by subtle
but irrelevant features. Note this kind of generalization is not guaranteed just by using a neural network; for example, Table~\ref{tb:test}
shows networks trained on normal quality JPEG images are confounded by low quality JPEG test images.

\begin{figure}[h]
\begin{center}
\resizebox{0.48\textwidth}{!}
{
\includegraphics{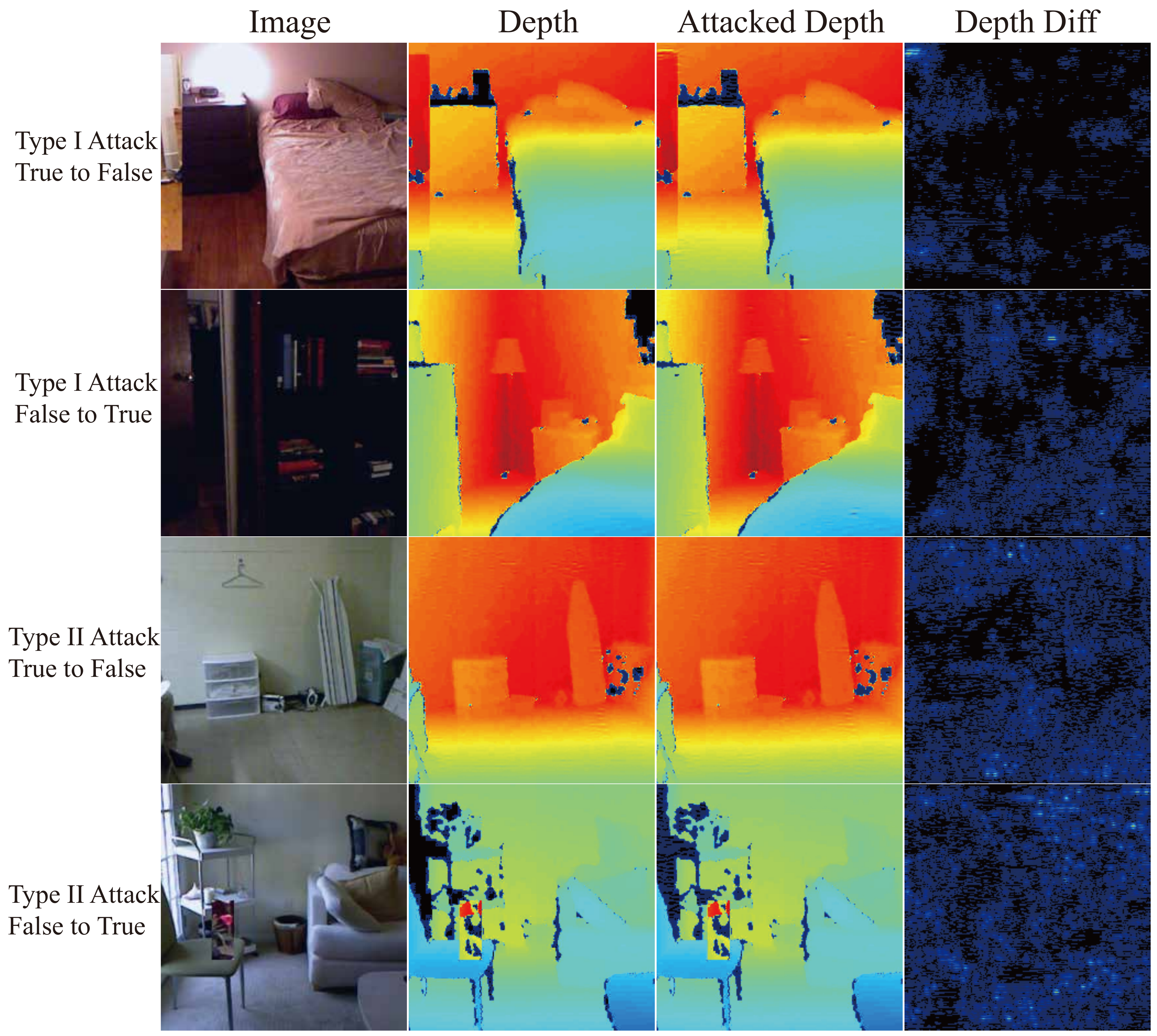}
}
\caption{We show figures for successful Type I attacks (fool the classifier) on the original classifier network, and successful Type II attacks (fool both
the classifier and detector) on our SafetyNet. Attackers are only allowed to manipulate the depth. Our SafetyNet is very difficult to attack and
 attacks changing label from False to True is harder. Successful attacks on our SafetyNet requires the original inputs hard to classify and the attacks
also need to manipulate the images more. }
\label{typeattack}
\end{center}
\vspace{-2ex}
\end{figure}

\begin{table*}[h!]
\begin{center}
\resizebox{0.9\textwidth}{!}
{
\begin{tabular}{  c | c | c | c | c | c | c | c | c | c}
 & \multicolumn{3}{c|}{Non Attack} & \multicolumn{3}{c|}{Type I Attack} & \multicolumn{3}{c}{Type II Attack} \\
  \hline
Method & F$\rightarrow$T & T$\rightarrow$F & T$\rightarrow$T reject & F$\rightarrow$T & T$\rightarrow$F & T$\rightarrow$T reject & F$\rightarrow$T & T$\rightarrow$F & T$\rightarrow$T reject \\
  \hline
Non Attack Data & 9.7\% & 0\% & 9.4\% & N/A & N/A & N/A & N/A & N/A & N/A\\
Unfamiliar Data Average & 17.3\% & 0\% & 0\% & N/A & N/A & N/A & N/A & N/A & N/A\\
Gradient Descent Attack & N/A & N/A & N/A & 9.9\% &  5.0\% & 6.1\% & 16.3\% & 3.7\% & 6.2\%\\
Transfer Attack Average & N/A & N/A & N/A &  4.6\% &  9.4\% & 33.6\% & 7.9\% &  9.8\% & 26.6\% \\
\end{tabular}
}
\vspace{-2ex}
\caption{Summary of our fc7 RBF-SVM detector's reaction on various non attack data and Type I, Type II attacks (smaller is better). 
\textbf{F$\rightarrow$T} means the rate at which false label images are classified as true and the detector does not spot, same for \textbf{T$\rightarrow$F}. 
\textbf{T$\rightarrow$T reject} means the rate at which true label images are classified as true, however,
they are rejected by the detector. This number only matters for non attack data because attacks are likely to distort activation patterns
even when the label keeps same. As expected, Type I attacks are less successful than Type II attacks. This is because a Type I attack does not explicitly
try to fool the detector. }
\label{tb:summary}
\end{center}
\end{table*}

\section{Rejecting by Classification Confidence}
Our experiments demonstrate that there is a trade-off between classification confidence and detection easiness for adversarial examples. 
Adversarial examples with high confidence in wrong classification labels tend to have more abnormal activation patterns, so they are easier to be
detected by detectors. While adversarial examples with low classification confidence in wrong labels are harder to be detected. For example,
attacks like DeepFool add small and just enough perturbations to change the classification label, so these adversarial examples are sometimes hard to detect. 
However, these adversarial examples could not assign high classification confidence to the wrong label. If they perform more iterations and increase
the wrong class classification confidence, our detector could detect them much easier. 

Experiments also show that Type II attacks on our quantized SVM detector together with the classifier produce adversarial examples with low confidence. All these experiments
mean that we can use classification confidence as a detection criteria, and it could help us increase the detector's detection ability and decrease the potential
to be attacked by Type II attacks. 

The classification confidence in our experiments is measured by the ratio of the example's second highest classification confidence to the highest classification confidence. 
For example, if an image has 60\% probability to be a dog and 15\% probability to be a cat, our classification confidence is 0.25. 
We reject examples with classification confidence ratio bigger than a threshold, which means the classifier is unsure about the classification. 

The classification confidence rejection results for non attack images and various Type II attack adversarial examples are included in 
Table~\ref{tb:rej_cifar} for Cifar-10 and Table~\ref{tb:rej_imagenet} for ImageNet-1000. Both tables show that rejecting by classification confidence rejects few non 
attack images while hugely increase the rejection of Type II attack adversarial examples. The benefits of rejecting by classification confidence is also demonstrated in
the Type II attacks section. 

\begin{table*}[h!]
\begin{center}
\resizebox{0.75\textwidth}{!}
{
\begin{tabular}{  c | c | c | c | c | c | c }
 & Statistics & Non Attack & L0 (II) & L2 (II) & Fast (II) & DeepFool (II) \\
\hline
	& Mean-confident & 95.45\% & 73.95\% & 69.36\% & 74.73\% & 73.71\% \\
m-SVM Det & Mean-ratio & 0.05 & 0.29 & 0.36 & 0.31 & 0.36 \\
	& Rejection-rate & 7.22\% & 43.58\% & 53.96\% & 45.46\% & 63.22\% \\
\hline
	& Mean-confident & 95.45\% & 95.71\% & 96.68\% & 79.21\% & 73.72\% \\
Subnet Det	& Mean-ratio & 0.05 & 0.03 & 0.04 & 0.25 & 0.36 \\
	& Rejection-rate & 7.22\% & 3.98\% & 5.50\% & 37.73\% & 63.22\% \\
\end{tabular}
}
\caption{CIFAR-10 classification confidence rejection results on non attack images, and various gradient descent based Type II attack adversarial examples. 
\textbf{Mean-confident} is the mean of classification confidence for the label with highest probability.
\textbf{Mean-ratio} is the mean of the ratio of the second highest predicted label confidence to the highest predicted label confidence.
\textbf{Rejection-rate} is the rate that examples are rejected because the ratio is higher than the threshold. The ratio for Cifar-10 is 0.25, which means the first predicted label 
confidence must be four times higher than the second one. For non attack data, the classification confidence rejection only rejects small amount of examples; for  quantized SVM 
detector, it rejects majority of Type II attack adversarial examples; for detection subnetwork, the rejection is not as efficient as quantized SVM detector, because getting high 
classification confidence while fooling detection subnetwork is easier (compared to quantized SVM detector). }
\label{tb:rej_cifar}
\end{center}
\vspace{-0.05in}
\end{table*}

\begin{table*}[h!]
\begin{center}
\resizebox{0.88\textwidth}{!}
{
\begin{tabular}{  c | c | c | c | c | c | c | c }
 & Statistics & Non Attack & L0 (II) & L2 (II) & Fast (II) & DeepFool (II) & DeepFool5 (II) \\
\hline
	 & Mean-confident & 81.55\% & 76.80\% & 41.25\% & 40.64\% & 43.93\% & 37.83\% \\
m-SVM Det & Mean-ratio & 0.15 & 0.17 & 0.43 & 0.49 & 0.77 & 0.51 \\
	 & Rejection-rate & 10.98\% & 14.26\% & 43.89\% & 49.55\% & 95.51\% & 51.90\% \\
\hline
	 & Mean-confident & 81.55\% & 67.53\% & 67.13\% & 36.65\% & 43.93\% & 37.82\% \\
Subnet Det & Mean-ratio & 0.15 & 0.28 & 0.30 & 0.51 & 0.77 & 0.51 \\
	 & Rejection-rate & 10.98\% & 25.21\% & 28.55\% & 51.80\% & 95.51\% & 51.84\% \\
\end{tabular}
}
\caption{ImageNet-1000 classification confidence rejection results on non attack images, and various gradient descent based Type II attack adversarial examples. 
The table arrangement is same to Table~\ref{tb:rej_cifar}, and DeepFool5 is top-5 DeepFool. The rejection ratio threshold is 0.5. For non attack data, the classification 
confidence rejection only rejects small 
amount of examples; for  quantized SVM  detector and detection subnetwork, they reject majority of Type II attack adversarial examples. }
\label{tb:rej_imagenet}
\end{center}
\end{table*}

\section{Type II Attacks fail}
A type II attack involves a search for an adversarial example that will be (a) mislabelled and (b) not detected. 
We perform the gradient descent based Type II attacks for Cifar-10 and ImageNet-1000 with SVM detector, and compare to detection subnetwork 
~\cite{metzen2017detecting}. Because the gradients of detection subnetwork are better formed, it should be easier to attack with Type II gradient descent attacks. 

In our experiments for Cifar-10 and ImageNet-1000, we use different gradient descent based Type II attacks (L0, L2, Fast, DeepFool and top-5 DeepFool) to attack 
the detector and classifier at the same time. In the main paper, gradient descent based Type II attacks on SceneProof dataset use L2 LBFGS method. 

The summary for Type II attacks on Cifar-10 could be found in Table~\ref{tb:summary1}. The numbers reported in the table are the percentages of adversarial examples that
are both misclassified and undetected (lower is better). Without classification confidence rejection, quantized SVM detector and detection subnetwork perform similar under
Type II attacks for L0, L2 and Fast methods, and quantized SVM detector performs significantly better under DeepFool Type II attacks. With classification confidence rejection,
quantized SVM detector is very hard to attack and performs better than detection subnetwork on almost all attacking methods. The classification confidence rejection increases at maximum
 $7\%$ false rejection on non attack images. The detailed percentages of Type II attacks on Cifar-10 could be found in Table~\ref{tb:attack1}. 

The summary for Type II attacks on ImageNet-1000 could be found in Table~\ref{tb:summary2}.The table arrangement is same to Table~\ref{tb:summary1}, and DeepFool5
is top-5 DeepFool attack. Quantized SVM detector consistently performs better than detection subnetwork for various attacking methods and for both with classification confidence
rejection and without. It's very difficult to perform Type II attacks on quantized SVM detector with rejection.  The classification confidence rejection increases at maximum $10\%$
false rejection on non attack images.  The detailed percentages of Type II attacks
on ImageNet-1000 could be found in Table~\ref{tb:attack2}. 

\begin{table*}[h!]
\begin{center}
\resizebox{0.53\textwidth}{!}
{
\begin{tabular}{  c | c | c | c | c }
Method & L0 (II) & L2 (II) & Fast (II) & DeepFool (II) \\
\hline
m-SVM Det & 19.73 & 18.70 & 6.86 & 22.01 \\
m-SVM Det - R & \textbf{9.86} & \textbf{7.32} & 3.41 & \textbf{8.32} \\
Subnet Det & 20.73 & 12.30 & 1.89 & 96.24 \\
Subnet Det - R & 19.69 & 11.57 & \textbf{1.19} & 35.39 \\
\end{tabular}
}
\caption{Percentages of CIFAR-10 Type II attack adversarial examples that are both misclassified and undetected, lower is better. - R means classification confidence rejection is used (rejection ratio is 0.25), otherwise only the detector is on duty. Without classification confidence rejection, quantized SVM detector and detection subnetwork perform similar under
Type II attacks for L0, L2 and Fast methods, and quantized SVM detector performs significantly better under DeepFool Type II attacks. With classification confidence rejection,
quantized SVM detector is hard to attack and performs better than detection subnetwork on almost all attacking methods. }
\label{tb:summary1}
\end{center}
\end{table*}

\begin{table*}[h!]
\begin{center}
\resizebox{0.65\textwidth}{!}
{
\begin{tabular}{  c | c | c | c | c | c }
Method & L0 (II) & L2 (II) & Fast (II) & DeepFool (II) & DeepFool5 (II)\\
\hline
m-SVM Det & 25.15 & 26.40 & 12.97 & 45.26 & 30.08 \\
m-SVM Det - R & \textbf{23.19} & \textbf{15.05} & \textbf{8.26} & \textbf{2.32} & \textbf{15.52} \\
Subnet Det & 70.52 & 36.43 & 21.25 & 100.00 & 42.24 \\
Subnet Det - R & 52.56 & 26.66 & 12.16 & 4.49 & 21.99 \\
\end{tabular}
}
\caption{Percentages of IMAGENET-1000 Type II attack adversarial examples that are misclassified and undetected, lower is better. - R means classification confidence rejection is used (rejection ratio is 0.5), otherwise only the detector is on duty. Quantized SVM detector consistently performs better than detection subnetwork for various attacking methods and for both with classification confidence rejection and without. It is difficult to perform Type II attacks on quantized SVM detector with rejection. }
\label{tb:summary2}
\end{center}
\end{table*}

\section{Application: SceneProof}
\label{sec:sceneproof}

SceneProof is a model application of our SafetyNet, because it would not work with a network that is subject to adversarial examples.
We would like Alice to be able to prove to Bob that her photo is real without the intervention of a team of experts, and
we'd like Bob to have high confidence in the proof.   This proof needs to operate at large scales (i.e. anyone could produce a proof while taking a picture), and automatically.  

Current best methods to identify fake images require careful analysis of vanishing points~\cite{farid},
illumination angles~\cite{o2012exposing}, and shadows~\cite{kee2013exposing} (reviews in~\cite{farid,farid1}).  
Such analyses are difficult to conduct at large scales or automatically. RGB image editing is easy, with very powerful
tools available. We construct a proof by capturing an RGBD image (easily accessible with consumer depth sensors), 
which changes the security aspect because it's quite hard to edit a depth map convincingly and those edits need to be consistent with the image. The
proof of realness is achieved by a classifier that checks both image and depth and determines whether they are
consistent.  Such a system works if (a) the classifier is acceptably accurate (i.e. it can determine whether the pair is
real or not accurately); (b) it can detect a variety of adversarial manipulations of depth or image or both (i.e. type I
attacks fail) ; and (c) type II attacks generally fail.   We achieve this by using the SafetyNet architecture.

We are mainly concerned with attacks label ``fake'' images ``real''.  Natural attacks on our system
are: produce a depth map for an RGB image using some regression method to obtain an RGBD image (regression); 
manipulate RGBD image by inserting new objects; 
take an RGBD image labeled ``fake'' and manipulate it to be labeled ``real'' (type I adversarial);  
take an RGBD image labeled ``fake'' and manipulate it to be labeled ``real'' in a way that fools SafetyNet's adversary
detector (type II adversarial).   There is a wide range of available regression/adversarial attacks, and our system needs
to be robust to various methods which might be used to prepare the regression/adversarial attack.  

Real test data is easily obtained.  We use the raw Kinect captures of LivingRoom and Bedroom from NYU v2 dataset~\cite{Silberman:ECCV12}.  
However, fake data requires care.  To evaluate generalization over
different attacks, we omit some ``regression'' methods from the training data and use them only in test.
``Regression'' methods used in both train and test are: random swaps of depth and image planes; single image predicted depth~\cite{eigen2014depth}; 
rectangle cropped region insertion and random shifted or scaled misaligned depth and image. 
``Regression'' methods used {\em only} in test are:  all zero depth values; nearest neighbor down-sample and up-sampled images
and depths;  low quality JPEG compressed images and depths;  Middlebury stereo RGBD 
dataset~\cite{scharstein2014high} and Sintel RGBD dataset~\cite{Butler:ECCV:2012}(which should be classified ``fake'' because they are renderings). 
Refer to Figure~\ref{typeattack} for dataset and attacks. 

\begin{table*}[h!]
\begin{center}
\resizebox{0.72\textwidth}{!}
{
\begin{tabular}{  c | c  || c | c | c | c | c }
Test example type  & Classifier Acc & B & A & B $\wedge$ A & A$\vee$B & A$\vee$B, T\\
  \hline
Natural RGBD,  False & 91.8\% & 15.2\% & 17.1\% & 14.3\% & 18.8\% & 19.6\% \\
Natural RGBD, True & 97.7\% & 10.1\% & 11.6\% & 9.2\% & 12.7\% & 10.8\% \\
Adversarial RGBD, False & 33.1\% & 89.1\% & 88.6\% & 87.3\% & 90.4\% & 88.9\% \\
Adversarial RGBD, True & 15.3\% & 81.3\% & 81.0\% & 79.1\% & 83.3\% & 83.7\% \\
\end{tabular}
}
\vspace{0.5ex}
\caption{This table shows detailed information about behavior of classifiers and detectors, summarized in Table~\ref{tb:summary}. 
In this table, results are calculated on the whole SceneProof dataset testing split. 
Type I attacks on SceneProof using a familiar adversary (i.e. one used to train the detector) fail.  We report
  results for two detectors: A is applied to fc7 of VGG19; B is applied
  to fc6 of VGG19.  Each detector column gives the rate at which the
  detector labels examples as adversarial.  A$\vee$B  means the example is
  reported adversarial if either detector says it is adversarial.  A$\wedge$B means the example is
  reported adversarial if both detectors say it is
  adversarial. A$\vee$B, T gives the fraction of examples that
  were detected as adversarial among these that were labelled True by the classifier.  
We break out results by type and ground truth label. For example,  for natural RGBD images that do not
  represent real scenes (first row), the classifier labels 91.8\% correctly; of the 8.2\% incorrectly labelled true, the
A$\vee$B  adversary detector rejects 19.6\% as adversarial examples (last column). 
}
\label{familiaradv}
\vspace{-2ex}
\end{center}
\end{table*}

\begin{table*}[h!]
\begin{center}
\resizebox{0.72\textwidth}{!}
{
\begin{tabular}{  c | c  || c | c | c | c | c }
  Test example type  & Classifier Acc & B & A & B $\wedge$ A & A$\vee$B & A$\vee$B,T\\
  \hline
zero D channel  & 76.5\% & 6.5\% & 25.6\% & 6.1\% & 26.0\% & 82.0\% \\
down-up sampled  & 75.2\% & 54.9\% & 60.6\% & 51.3\% & 63.4\% & 87.6\% \\
low quality JPEG & 36.4\% & 80.1\% & 79.2\% & 77.2\% & 82.2\% & 81.8\% \\
Sintel RGBD~\cite{Butler:ECCV:2012} &  27.6\% & 45.3\% & 51.7\% & 39.7\% & 57.2\% & 61.4\% \\
Middlebury RGBD~\cite{scharstein2014high} &  24.0\% & 39.7\% & 40.3\% & 33.4\% & 46.6\% & 47.8\% \\
\end{tabular}
}
\vspace{0.5ex}
\caption{This table shows detailed information about behavior of classifiers and detectors, summarized in Table \protect~\ref{tb:summary}.
The table arrangement is same to Table~\ref{familiaradv}.
Type I attacks on SceneProof using an unfamiliar adversary (i.e. one not used to train the detector) generally fail. 
All these examples should be labelled false, {\em OR} rejected as adversarial.  The column for each detector reports the rate at
  which the detector identifies examples as adversarial.  For example, in the first row, 76.5\% of zero D channel
  RGBD images are correctly labelled as  false by the classifier; of those labelled ``true'', 82.0\% are rejected as
  adversarial (last column).  This means that a  total of 4.2\% of zero D channel RGBD images pass through SafetyNet with ``true'' labels.
}
\label{tb:test}
\vspace{-2ex}
\end{center}
\end{table*}

{\bf Type I attacks on SafetyNet fail:} Type I attacks on SceneProof using a familiar adversary (i.e. one used to train the detector) fail.  We report
  results for two detectors A (applied to fc7 of VGG19) and B (applied to fc6 of VGG19) in Table~\ref{familiaradv}.
  Type I attacks on SceneProof using an unfamiliar adversary (i.e. one not used to train the detector) generally   fail. We report
  results for two detectors A (applied to fc7 of VGG19) and B (applied to fc6 of VGG19) in Table~\ref{tb:test}. 

A type II attack  must both fool the classifier {\em and} sneak past the detector.  We distinguish between 
two conditions. In non-blackbox case, the internals of the SafetyNet system is {\em accessible} to the attacker.  Alternatively, the
network may be a black box, with internal states and gradients concealed. In this case, attackers must probe with
inputs and gather outputs, or build another approximate network as in~\cite{papernot2016practical}.

{\bf Type II attacks on accessible SafetyNet fail:} a type II attack involves a 
 search for an adversarial example that will be (a)
mislabelled and (b) not detected.  This search is made difficult by the quantization procedure and by the narrow basis
functions in the RBF-SVM, so we smooth the quantization operation and the  RBF-SVM kernel operation.
Smoothing is essential to make the search tractable, but can significantly misapproximate SafetyNet (which is what makes
attacks hard).    Our smoothing attack uses a sigmoid function with 
parameter $\lambda$ to simulate the quantization process.  We also help the search process by increasing the size of the RBF parameter $\sigma$ to
form smoother gradients.  Even after smoothing the objective function, attacks tend to fail, likely because it is hard to make an effective tradeoff between easy
search and approximation. Table~\ref{tb:attack} includes Type I and Type II, blackbox and non-blackbox attacking results on SceneProof dataset. Our SafetyNet
is the most robust architecture to various attacks. 

\textbf{Type II attacks on black box SafetyNet fail:} Assume the state of SafetyNet is concealed.  We follow~\cite{papernot2016transferability,
  moosavi2016universal} by building attacks on various alternative networks, then transferring these network's adversarial samples. These attacks fail for
our SafetyNet, refer to Table~\ref{tb:attack}. In contrast to SafetyNet, the detector subnetwork of~\cite{metzen2017detecting} is generally susceptible 
to type II attacks in both blackbox and non-blackbox settings. This is because of quantization process and detection subnetwork's classification boundary 
problem~\cite{moosavi2016universal}.

\begin{table*}[h!]
\begin{center}
\resizebox{1.0\textwidth}{!}
{
\begin{tabular}{  c | c | c || c | c | c | c | c | c | c | c | c }
  Method & \multicolumn{2}{c||}{Ori} & \multicolumn{3}{c|}{Subnet Det} & \multicolumn{3}{c|}{Det A} & \multicolumn{3}{c}{Det ABC} \\
  \hline
 & F$\rightarrow$T & T$\rightarrow$F & F$\rightarrow$T & T$\rightarrow$F & T$\rightarrow$T reject & F$\rightarrow$T & T$\rightarrow$F & T$\rightarrow$T reject & F$\rightarrow$T & T$\rightarrow$F & T$\rightarrow$T reject\\
  \hline
 Non Attack Data & 16.3\% & 0.6\% & \textbf{8.4\%} & \textbf{0\%} & 10.2\% & 9.7\% & \textbf{0\%} & \textbf{9.4\%} & \textbf{8.4\%} & \textbf{0\%} & 9.9\%\\
  \hline
 Gradient Descent (I)& 32.8\% & 55.3\% & 13.4\% & 9.5\% & 6.0\% & 9.9\% & 5.0\% & 6.1\% & \textbf{8.4\%} & \textbf{0.3\%} & 6.3\% \\
 VGG FastSign TF  (I) & 30.6\% & 2.8\% & 14.9\% & 2.2\% & 54.1\% & 7.5\% & 2.5\% & 44.1\% & \textbf{6.6\%} & \textbf{1.9\%} & 47.2\%\\
 ResNet GradDesc TF (I) & 28.9\%  & 36.7\% & 15.3\% & 22.4\% & 33.2\% & 3.6\% & 13.4\% & 29.1\% & \textbf{2.7\%} & \textbf{11.9\%} & 30.3\%\\
 ResNet FastSign TF (I) & 22.2\% & 29.1\% & 7.6\% & 15.1\% & 29.8\% & 2.8\% & 12.2\% & 27.5\% & \textbf{2.2\%} & \textbf{11.6\%} & 27.8\% \\
 Type I Average & 28.6\% & 30.9\% & 12.8\% & 12.3\% & 30.8\% & 6.0\% & 8.3\% & 26.7\% & \textbf{5.0\%} & \textbf{6.4\%} & 27.9\% \\
  \hline
 Gradient Descent (II)& 32.8\% & 55.3\% & 26.3\% & 21.9\% & 11.9\% & 16.3\% & 3.7\% & 6.2\% & \textbf{13.2\%} & \textbf{2.6\%} & 9.6\% \\
 VGG Finetune TF (II) & 20\% & 3.1\% & \textbf{17.1}\% & \textbf{0\%} & 43.5\% & 17.2\% & \textbf{0\%} & 45.6\% & 17.2\% & \textbf{0\%} & 48.4\% \\
VGG Subnet Det TF (II) &  16.3\% & 0.6\% & 13.7\% & \textbf{0\%} & 15.6\% & 10.3\% & \textbf{0\%} & 12.5\% & \textbf{9.1\%} & \textbf{0\%} & 13.1\%\\
 ResNet Finetune TF (II) & 15.6\% & 40.3\% & 8.5\% & 31.3\% & 29.3\% & 1.3\% & 27.2\% & 20.6\% & \textbf{0.3\%} & \textbf{25\%}  & 21.0\% \\
 ResNet Subnet Det TF (II) & 23.8\% & 29.7\% & 17.6\% & 19.3\% & 29.8\% & 2.8\% & 12.2\% & 27.5\% & \textbf{2.2\%} & \textbf{11.6\%} & 27.5\%\\
 Type II Average & 21.7\% & 25.8\% & 16.6\% & 14.5\% & 26.0\% & 9.6\% & 8.6\% & 22.5\% & \textbf{8.4\%} & \textbf{7.84\%} & 23.9\% \\
\end{tabular}
}
\vspace{1.0ex}
\caption{Type I and Type II attacks, non-blackbox and blackbox attacks on SceneProof all fail. This table is gather by attacking a randomly selected subset of 
3200 images from the whole SceneProof dataset test split (contains 80K images). The table compares a VGG19 network (Ori) with 
the detection subnetwork of \protect \cite{metzen2017detecting} (Subnet), and two variants of SafetyNet (Det A, where we
have an RBF-SVM on fc7; and Det ABC, where we have an RBF-SVM on each of fc7, fc6 and pool5, and declare an adversary when
any detector responds). T$\rightarrow$F shows the rate at which true label classified as false and not detected and F$\rightarrow$T  shows false 
label classified as true and not detected (i.e. lower is better). T$\rightarrow$T reject shows the rate at which true samples are classified as true, but rejected by 
detector. \emph{This rate only matters for non attack data, and does not matter for all attacks because attacks are likely to distort the activation patterns even 
if the classification label has not been changed.} There is no manipulation for the non attack data, which
represents unforced errors by the classifier; note that each of the adversary detectors catches a high percentage of the
false positives committed by the classifier and rejects them as adversarial. We group attacks by type I attack (I) and type II attack (II). 
The gradient descent shows the performance of an attack by gradient descent method (type I or type II) on an accessible network. Even 
when the network is accessible, attacks tend to be unsuccessful. TF represents
blackbox transfer attacks where adversarial samples are obtrained from another network (VGG - a VGG19 model; ResNet - a ResNet model). 
The VGG19 (ResNet) FastSign TF gives results for a type I attack by transferring FastSign adversarials from a VGG19 (ResNet) model. 
VGG (ResNet) Finetune TF finetunes a VGG19 (ResNet) network with adversarial examples labelled false, and generate adversarials;
VGG (ResNet) Subnet Det TF uses a VGG19 (ResNet) network with the detection subnetwork of~\cite{metzen2017detecting}. 
The results show that original classifier 
network is easy to attack successfully with all attacking methods. Subnet methods can detect type I attacks, but are not robust to transfer 
attacks and are vulnerable to type II attacks. Our SafetyNet is robust to Type I and Type II attacks, as well as gradient descent and transfer 
attacks, likely because: quantization hides irrelevant patterns; SafetyNet works like a matcher, so is hard to differentiate; and the subnetwork 
suffers from the classification boundary problem noted in~\cite{moosavi2016universal}.
}
\label{tb:attack}
\vspace{-5ex}
\end{center}
\end{table*}

\section{Theory: Bars and P-domains}
\label{sec:label}

We construct one possible explanation for adversarial examples that successfully explains (a) the phenomenology and (b)
why SafetyNet works. In this explanation, we assume the network uses ReLU and weight decay, because they are representative, make
it easier to explain, and likely to extend to other conditions with some modifications. We have a network with $N$ layers of ReLU's, and 
study $y_i^{(k)}(\vect{x})$, the values at the output of the $k$'th layer of ReLUs.  
This is a piecewise linear function of $\vect{x}$.  Such functions break up the input space into {\em cells}, at whose boundaries  the
piecewise linear function changes (i.e. is only $C^0$).  Now assume that for some $y^{(k)}_i(\vect{x})$ there exist  {\em p-domains}
(union of cells) ${\cal D}$ in the input space such that: (a) there are no or few examples in the p-domain; (b) the
measure of ${\cal D}$ under $P(X)$ is small; (c) $\dafabs{y^{(k)}_i(\vect{x})}$ is large inside ${\cal D}$ and small
outside ${\cal D}$.  We will always use the term ``p-domain'' to refer to domains with these properties.  We think that 
the total measure of all p-domains under $P(X)$ is small.  

\begin{figure}[h]
\centerline{
\resizebox{0.6\columnwidth}{!}
{
\centerline{\includegraphics{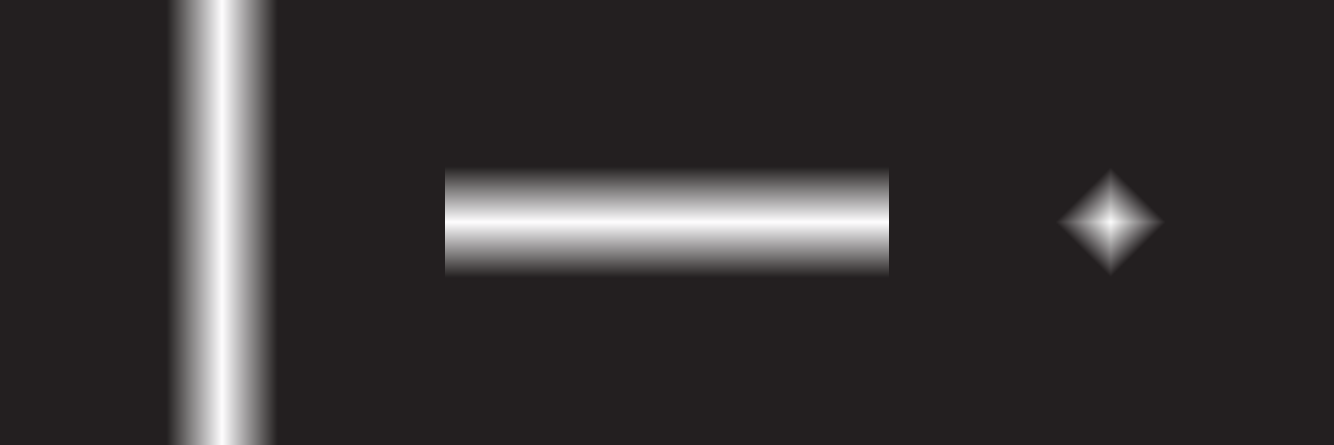}}
}}
\caption{Simple example bar functions on the $x$, $y$ plane, where black is 0 and white is 1.  {\bf Left:}
  $\phi(\vect{x}; 1, 0, 1)$ (i.e. a bar in $x$, independent of $y$); {\bf center:}  $\phi(\vect{x}; 2, 0, 1)$; and {\bf
    right:} $b(\vect{x}; \left\{1, 2\right\}, \vect{0}, 1)$.}
\label{figbar}
\vspace{-2ex}
\end{figure}

By construction, ReLU networks can represent such p-domains. We construct a p-domain using a basis function with small support. 
$\relu{\vect{u}}$ denote a ReLU applied to $\vect{u}$.  We have {\em basic bar function} $\phi$.
\[
\phi(\vect{x}; i, s, \epsilon)=\frac{1}{\epsilon}\left(\begin{array}{c}\relu{(x_i-s)+\epsilon}-\\2\relu{(x_i-s)}+\\\relu{(x_i-s)-\epsilon}\end{array}\right)
\]
where $\phi$ has support when $\dafabs{x_i-s}<\epsilon$ and has peak value $1$. For an index set ${\cal I}$ with cardinality $\#{\cal I}$ and vectors $\vect{s}$, $\epsilon$, we write {\em bar function} $b$ as
\[
b(\vect{x}; {\cal I}, \vect{s}, \epsilon)=\relu{\left(\sum_{i \in {\cal I}} \phi(\vect{x}; i, s_i, \epsilon_i) - \#{\cal I}+1\right)}
\]
where $b$ has support when $\lonenorm{\vect{x}_{\cal I}-\vect{s}_{\cal I}}<\epsilon_{\cal I}$. Figure~\ref{figbar} illustrates these functions.  
It is clear that a CNN can encode bars and weighted sums of bars, and that for at least $k\geq 2$ every $y_i^{(k)}$ could in principle be a bar function.  
Appropriate choices of $\vect{s}$, $\epsilon$ and ${\cal I}$ choose the location and support of the bar and so can produce bars which have low measure under $P(X)$.   
Now the functions presented to the softmax layer are a linear combination of the $y_i^{(N)}(\vect{x})$.  This means that with choice of weight and parameters, 
a bar can appear at this level, and create a p-domain.  

We expect such p-domains to have several important properties.  
{\bf Adversarial fertility:} P-domains can be used to make adversarial examples by choosing a  point
in a p-domain close to $\vect{x}$.  Because there are no or few examples in the p-domain, the loss may not cause the classifier  to 
control the maximum value attained by $y_i^{(k)}(\vect{x})$ in this p-domain; and the large range of values inside the p-domain can be used to change  the 
values in layers upstream of $k$, by moving the example around the p-domain. 
{\bf Generalization-neutral:}  The requirement that p-domains have small measure in $P(X)$ means that both train and test
examples are highly unlikely to lie in   p-domains.  A system with p-domains could generalize well without being immune to
adversarial examples. 
Some subset of p-domains are likely {\bf findable by LBFGS.} Consider the gradient of  $y_i^{(N)}(\vect{x})$ with respect to $\vect{x}$ in two cells separated by 
a boundary, where some ReLU changes state,  weight decay encourages a relatively small change in gradient over these boundaries. 
If cells neighboring a p-domain have no or few examples in them, 
we can expect that the gradient change within cell is small too and a second order approximation of $y_i^{(N)}(\vect{x})$ could be reliable. 
We also expect cells to be small, so
search and entering a p-domain are possible and  requires crossing multiple cell boundaries, which means
many changes in ReLU activation.  This argument suggests p-domains present {\bf odd patterns of ReLU activation},  particularly in p-domains where some of the
$y_i^{(k)}(\vect{x})$ are large in the absence of examples.

{\bf Why p-domains could exist:} As Zhang {\em et al.} point out, the number of training examples available to a typical modern
network is small compared to the relative capacity of deep networks~\cite{zhang2017general}.  For example, excellent training error is obtainable for randomly chosen image labels~\cite{zhang2017general}.
We expect that $y_i^{(N)}(\vect{x})$ will have a number of cells that is exponential in the dimension of $\vect{x}$, ensuring that the vast majority of cells lack any
example.  However, the weight decay term is not sufficient to ensure that $y_i^{(N)}$ is zero in these cells.  Overshoot by stochastic gradient descent, caused by poor scaling in the loss, is the likely reason that 
$y_i^{(N)}(\vect{x})$ has support in these cells.  Szegedy {\em et al.} demonstrate that, in practice, ReLU layers can
have large norm as linear operators, despite weight decay (see~\cite{szegedy2013intriguing}, sec. 4.3), so large
values in p-domains are plausible.  This large norm is likely to be the result of overshoot.  Recall that the value of
$y_i^{(N)}(\vect{x})$ is  determined by the {\em product} of numerous weights, so in some locations in
$\vect{x}$, the value of $y_i^{(N)}$ could be  large, which is a result of multiple layer norms interacting poorly.

An alternative to attacking by search using smoothed RBF gradients is as follows. One might pass an example through the
main classifier, determine what code it had, then seek an adversarial example that produces that code (and so must fool
the RBF-SVM).  We sketch a proof that the optimization problem is extremely difficult.
Choose some threshold $t>0$.  We use $b_t(u)$ for the function that binarizes its argument with $t$.  Assume we have at least one unit $y_i^{(k)}$ that encodes
a weighted sum of bar functions.  We wish to create an adversarial example $\vect{a}(\vect{x}^*)$ that (a) meets criteria for being adversarial and (b) ensures 
that $b_t(y_i^{(k)}(\vect{a}))$ takes a prescribed value (either one or zero).   The feasible set for this constraint can be disconnected (\eg a sum of the bump functions of Figure~\ref{figbar} (right)), and so need not be 
convex, implying that the optimization problem is intractable.  As a simple example, the following constraint set is disconnected for $\epsilon < 1/2$ 
\[
\left\{ \vect{x} \mid b_t(b(\vect{x}; 1, \vect{0}, \epsilon)+b(\vect{x}; 1, \vect{1}, \epsilon))=1\right\}.
\]

\section{Discussion}

We have described a method to produce a classifier that identifies and rejects adversarial examples. Our SafetyNet is
able to reject adversarial examples that come from attacking methods not seen in training data. We have shown that it is hard to
produce an example that (a) is mislabeled and (b) is not detected as adversarial by SafetyNet.  We have sketched one
possible reason that SafetyNet works, and is hard to attack. Many interesting problems are opened by our work, and we provides 
lots of insights into the mechanism that neural network works.

\textbf{SaferNet:} There might be some better architecture than our SafetyNet, whose objective function is harder to optimize. 
The ideal case would be an architecture that forces the attacker to solve a hard discrete optimization problem which does 
not naturally admit smoothing. 

\textbf{Neural network pruning: } Our work suggests that networks behave poorly for input space regions
where no data has been seen.  We speculate that this behavior could be discouraged by a post-training pruning process,
which removes neurons, paths or activation patterns not touched by training data. 

\textbf{Explicit management of overshoot during training:} we have explained adversarial examples using p-domains, which
is the result of poor damping of weights during training.  We speculate that constructing
adversarial examples during training, by identifying locations where this damping problem occurs and exploiting structural insights 
into network behavior, could control the adversarial sample problem (rather than just using adversarial examples as training data).

\section{Acknowledgements}
This work is supported in part by ONR MURI Award N00014-16- 1-2007,  in part by NSF under Grant No. NSF IIS- 1421521, and in part by a Google MURA award.

\section{Supporting Materials}
\subsection{SceneProof Dataset}
\begin{figure*}[h]
\begin{center}
\resizebox{0.9\textwidth}{!}
{
\includegraphics{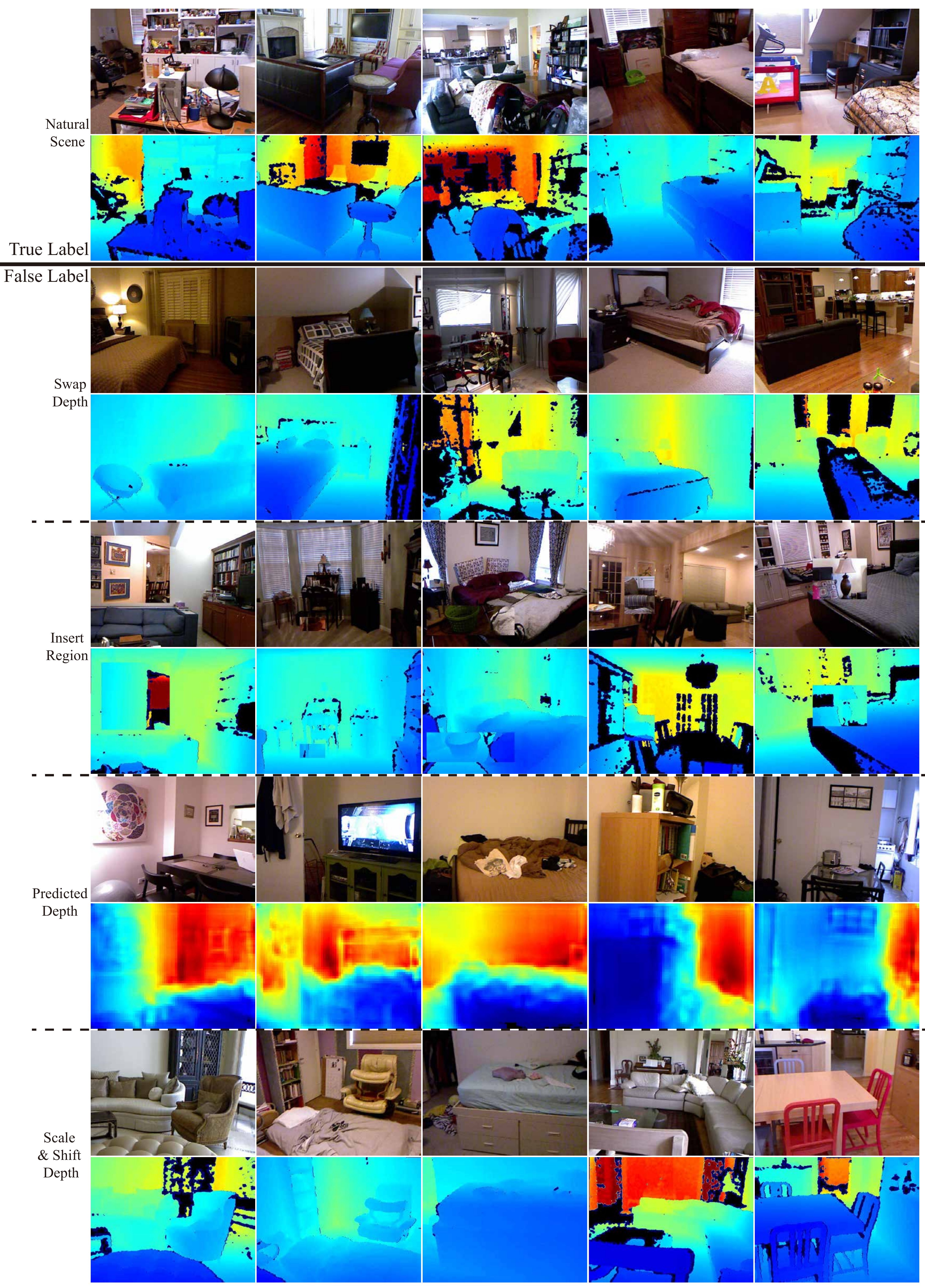}
}
\caption{SceneProof dataset part I. Natural Scene has true label, and others have false labels. }
\label{dataset}
\end{center}
\vspace{-2ex}
\end{figure*}

\begin{figure*}[h]
\begin{center}
\resizebox{0.9\textwidth}{!}
{
\includegraphics{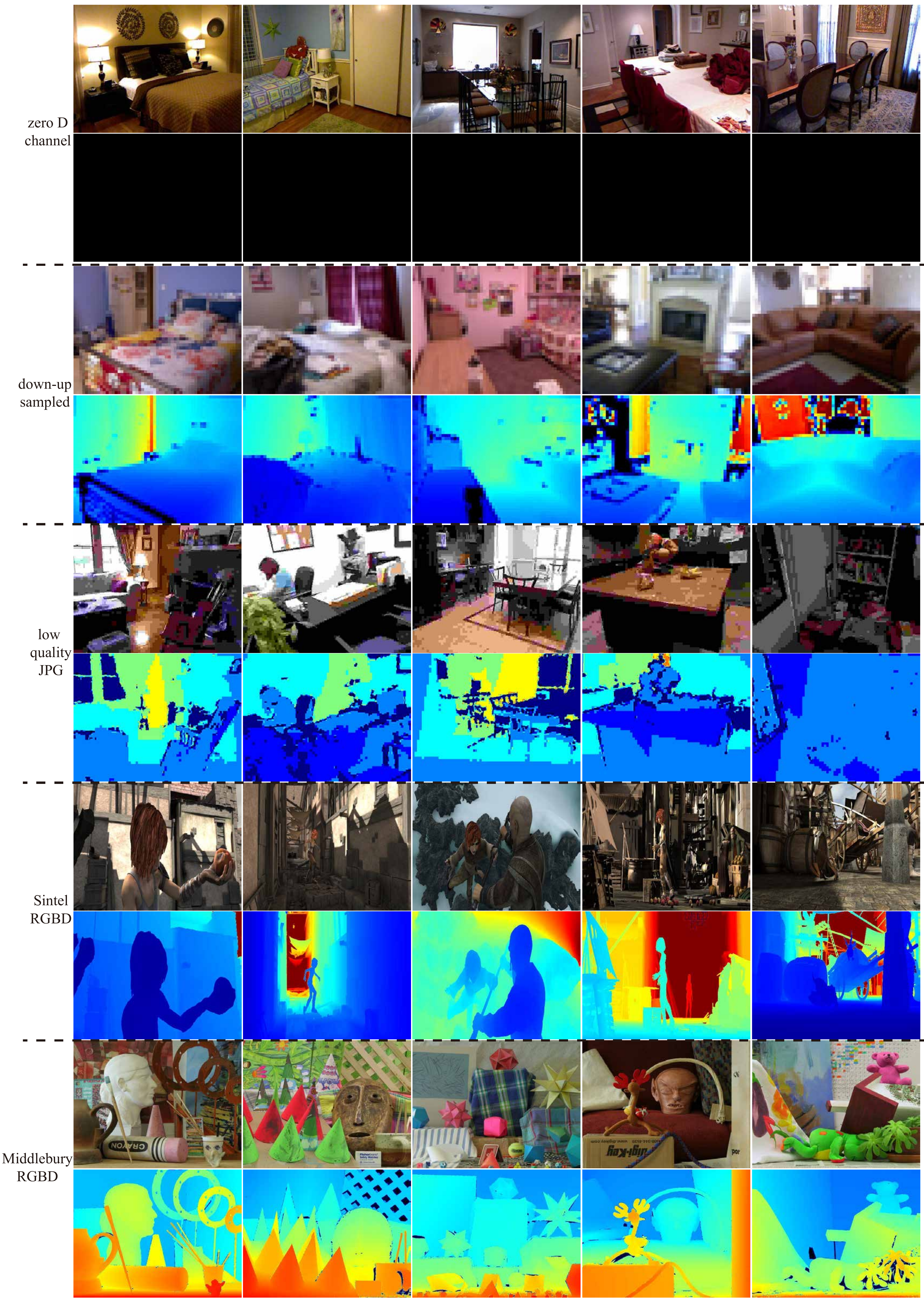}
}
\caption{SceneProof dataset part II. All have false labels.}
\label{dataset2}
\end{center}
\vspace{-2ex}
\end{figure*}

Our SceneProof dataset is processed from NYU Depth v2 raw captures, Sintel Synthetic RGBD dataset and Middlebury Stereo dataset. 
The dataset is split into part I and part II. Part I contains NYU natural image \& depth pairs, along with manipulated unnatural scenes (swap depth,
insert region, predicted depth, scale \& shift depth), refer to Figure~\ref{dataset}. It is used to train our classifier and work as test data part I. 
Part II contains unnatural scenes manipulated by other methods (set depth channel to zero, down sample and then up-sample both RGBD channels, 
aggressively compress the JPG RGBD images), and image \& depth pairs from synthetic dataset and stereo dataset, refer to Figure~\ref{dataset2}. 
Part II is used as test data part II to test the generalization ability of our SceneProof network, and check the reactions of
our detectors to unseen unnatural inputs. A good detector need to tend to reject unfamiliar data type, which does not exist in training data, because
it is hard for classifier to do right classifications on unseen data types. In real application scenarios, it needs to be a human computer hybrid system
where computer provides suspicious cases and human makes final decisions. Table~\ref{dataset_details} includes the dataset constitution, and we plan
to release the dataset for academia usages. 

\subsection{Type II Attacks on Cifar-10 and ImageNet-1000}
In this section, we include the detailed percentages of Type II attacks on Cifar-10 could be found in Table~\ref{tb:attack1}, and 
the detailed percentages of Type II attacks on ImageNet-1000 could be found in Table~\ref{tb:attack2}. 

\begin{table}[h]
\centering
{
\begin{tabular}{ c | c | c | c }
   & Training & Testing I & Testing II \\
  \hline
  Natural Scene & 141780 & 57542 & N/A\\
  Swap Depth & 33927 & 16094 & N/A\\
  Insert Region & 30426 & 13741 & N/A\\
  Predicted Depth & 53904 & 17026 & N/A\\
  Scale\&Shift Depth & 23523 & 10681 & N/A\\
  zeroD channel& N/A & N/A & 1449 \\
  down-up sampled& N/A & N/A & 1449 \\
  low quality JPG& N/A & N/A & 1449 \\
  Sintel RGBD& N/A & N/A & 54\\
  Middlebury RGBD& N/A & N/A & 30\\
 Total & 283560 & 115084 & 4431 \\
\end{tabular}
}
\vspace{2mm}
\caption{Number of image \& depth pairs for each data type in each dataset split. Natural Scene has true label and the other data types have false labels. }
\label{dataset_details}
\end{table}

\begin{table*}[h]
\begin{center}
\resizebox{0.75\textwidth}{!}
{
\begin{tabular}{c | c |c c| c c|c c|c c }
 & & \multicolumn{2}{c|}{L0 (II)} & \multicolumn{2}{c|}{L2 (II)} & \multicolumn{2}{c|}{Fast (II)} & \multicolumn{2}{c}{DeepFool (II)}\\
\hline
Cifar-10 & &   undet   &   det & undet   &   det  & undet   &   det  & undet   &   det \\
\hline
\hline
\multirow{2}{*}{m-SVM Det} & = & 37.95 & 22.58 &51.16 & 19.23 &33.45 & 41.75 & 1.03 &  2.71 \\
& $\not=$& 19.73 & 19.75 & 18.70 & 10.90 & 6.86 & 17.95 & 22.01 & 74.23\\
\hline
\multirow{2}{*}{m-SVM Det - R} & = &  31.79 & 28.74 &  42.23 & 28.17 &  30.87 & 44.32 &   0.41 &  3.34\\
 & $\not=$&  9.86 & 29.62 & 7.32 & 22.29 &  3.41 & 21.39 & 8.32 & 87.94\\
\hline

\multirow{2}{*}{Subnet Det} & = & 16.91 & 21.64 & 28.57 & 32.01 & 8.06 & 66.57 & 3.76 & 0.00 \\
 & $\not=$ & 20.73 & 40.72 & 12.30 & 27.13 & 1.89 & 23.48 & 96.24 & 0.00 \\
\hline
\multirow{2}{*}{Subnet Det - R} & = & 16.25 & 22.30 & 28.02 & 32.56 & 7.53 & 67.10 & 1.15 & 2.61 \\
 & $\not=$ & 19.69 & 41.76 & 11.57 & 27.85 & 1.19 & 24.18 & 35.39 & 60.85 \\

\end{tabular}
}
\caption{Percentage details of Table~\ref{tb:summary1} with correct classification (=) and undetected as adversarials (undet), correct classification and detected as adversarials (det), 
misclassification ($\not=$) and undetected as adversarials, misclassification and detected as adversarials. Table~\ref{tb:summary1} comes from misclassification and undetected as adversarials
(left down corner). \emph{For all Type II attacks, correct classification and detected as adversarials percentage
does not matter, because attacks tend to distort activation patterns even when the labels have not been changed.} }
\label{tb:attack1}
\end{center}
\end{table*}

\begin{table*}[h]
\begin{center}
\resizebox{0.9\textwidth}{!}
{
\begin{tabular}{c | c |c c| c c|c c|c c| c c }
 & & \multicolumn{2}{c|}{L0 (II)} & \multicolumn{2}{c|}{L2 (II)} & \multicolumn{2}{c|}{Fast (II)} & \multicolumn{2}{c|}{DeepFool (II)} & \multicolumn{2}{c}{DeepFool5 (II)}\\
\hline
ImageNet-1000 & &   undet   &   det & undet   &   det  & undet   &   det  & undet   &   det  & undet & det\\
\hline
\hline

\multirow{2}{*}{m-SVM Det} & = &  0.00 &  0.00 & 3.12  &  1.58 & 55.21  &  7.04 &  0.00 &  0.00 &  0.00 &  0.00 \\
& $\not=$ &  25.15 &  74.84 & 26.40  &  68.90 &  12.97 & 24.78  &  45.26 & 54.74  & 30.08  &  69.92 \\
\hline
\multirow{2}{*}{m-SVM Det - R} &  = &  0.00  & 0.00  &  2.43 &  2.27 &  53.06 &  9.19 & 0.00  &  0.00 &  0.00 &  0.00 \\
& $\not=$ &  23.19  &  76.80 &  15.05 & 80.24  &  8.26 & 29.48  & 2.32  & 97.67  & 15.52  &  84.48 \\
\hline

\multirow{2}{*}{m-SVM Det} & = &  17.67  & 4.13  & 33.13  & 20.69  & 21.96  & 13.28  &  0.00 &  0.00 &  0.00 &  0.00 \\
& $\not=$ &  70.52  & 7.68  & 36.43  & 9.74  &  21.25 & 43.52  &  100.00 & 0.00  &  42.24 & 57.76  \\
\hline
\multirow{2}{*}{m-SVM Det - R} &  = &  16.03  &  5.77 &  30.86 &  22.97 &  19.63 & 15.61  & 0.00  & 0.00  & 0.00  &  0.00 \\
& $\not=$ &  52.56  & 25.64  & 26.66  & 19.51  &  12.16 & 52.61  &  4.49 &  95.51 &  21.99 & 78.01  \\

\end{tabular}
}
\caption{Percentage details of Table~\ref{tb:summary2} with correct classification (=) and undetected as adversarials (undet), correct classification and detected as adversarials (det), 
misclassification ($\not=$) and undetected as adversarials, misclassification and detected as adversarials. Table~\ref{tb:summary2} comes from misclassification and undetected as adversarials
(left down corner). \emph{For all Type II attacks, correct classification and detected as adversarials percentage
does not matter, because attacks tend to distort activation patterns even when the labels have not been changed.} }
\label{tb:attack2}
\end{center}
\end{table*}

{\small
\bibliographystyle{ieee}
\bibliography{egbib}
}

\end{document}